\documentclass[conference]{IEEEtran}
\IEEEoverridecommandlockouts
\usepackage{cite}
\usepackage{amsmath,amssymb,amsfonts}
\usepackage{algorithmic}
\usepackage{graphicx}
\usepackage{textcomp}
\usepackage{xcolor}
\usepackage{marvosym}
\usepackage{ifsym}
\usepackage{stfloats}
\usepackage{subfigure}
\usepackage{algorithm}
\usepackage{array}
\usepackage{multirow}
\usepackage{makecell}
\usepackage{booktabs}

\usepackage{pifont} 

\def\BibTeX{{\rm B\kern-.05em{\sc i\kern-.025em b}\kern-.08em
    T\kern-.1667em\lower.7ex\hbox{E}\kern-.125emX}}
\begin{document}

\title{DA-PFL: Dynamic Affinity Aggregation for Personalized Federated Learning
\thanks{\textrm{\Letter} Corresponding author.}
}

\author{\IEEEauthorblockN{Xu Yang\textsuperscript{1}, Jiyuan Feng\textsuperscript{1}, Songyue Guo\textsuperscript{1}, Ye Wang\textsuperscript{1, 2}, Ye Ding\textsuperscript{3},Binxing Fang\textsuperscript{1}, and Qing Liao\textsuperscript{1 \textrm{\Letter}}}
\IEEEauthorblockA{\textsuperscript{1}\textit{Harbin Institute of Technology (Shenzhen), }Shenzhen, China \\
\textsuperscript{2}\textit{National University of Defense Technology, }Changsha, China \\
\textsuperscript{3}\textit{Dongguan University of Technology, }Dongguan, China \\
\{xuyang97, fengjy, guosongyue\}@stu.hit.edu.cn, ye.wang@nudt.edu.cn, dingye@dgut.edu.cn, \\ fangbx@cae.cn, liaoqing@hit.edu.cn}

}

\maketitle

\begin{abstract}
Personalized federated learning becomes a hot research topic that can learn a personalized learning model for each client. Existing personalized federated learning models prefer to aggregate similar clients with similar data distribution to improve the performance of learning models. However, similarity-based personalized federated learning methods may exacerbate the class imbalanced problem. In this paper, we propose a novel Dynamic Affinity-based Personalized Federated Learning model (DA-PFL) to alleviate the class imbalanced problem during federated learning. Specifically, we build an affinity metric from a complementary perspective to guide which clients should be aggregated. Then we design a dynamic aggregation strategy to dynamically aggregate clients based on the affinity metric in each round to reduce the class imbalanced risk. Extensive experiments show that the proposed DA-PFL model can significantly improve the accuracy of each client in three real-world datasets with state-of-the-art comparison methods. 

\end{abstract}


\section{Introduction}
Personalized Federated Learning (PFL) aims to train a well-performing personalized model for each client by aggregating the knowledge of other clients without sharing its local data.  
PFL attracts widespread attention due to its great success in image recognition \cite{surveytkde1,ChenTowards,oh2022fedbabu,luo2021no}, recommendation system\cite{muhammad2020fedfast,wang2021fppfrs,Recommendertkde,FedDSRtkde,SocialRecommendertkde}, keyboard suggestion\cite{hard2018federated,SinghalFederated} and healthcare\cite{yang2021flop,dayan2021federated,chen2022pflBench} under Non-IID distribution in the real world. 

How to select and aggregate valuable clients based on similar data distribution to learn personalized models is a key issue of PFL. 
For example, Ghosh et al.\cite{ifcaghosh2020efficient} propose the Iterative Federated Clustering Algorithm (IFCA), which iteratively divides clients into different combinations by the similarity of samples and trains a model for the same clients. Ruan et al.\cite{ruan2022fedsoft} propose the FedSoft method, in which each client trains the personalized model and aggregates models using different combinations of the data distributions from other similar clients in each round. Huang et al.\cite{huang2021personalized} propose the FedAMP method, which encourages pairwise collaboration between clients with similar data by adopting an attention message-passing mechanism. Liu et al.\cite{liu2021pfa} propose the PFA method, which uses Euclidean Distance \cite{danielsson1980euclidean} to measure the similarity between data distributions. The above personalized federated learning methods mainly focus on selecting similar clients to train personalized models collaboratively \cite{ma2022layer,jeong2022factorized,ouyang2021clusterfl,palihawadana2022fedsim,vahidian2023efficient}.
\begin{figure}[!t]
\centering
\subfigure[Data distribution of four aggregated groups.]{\includegraphics[scale=0.38]{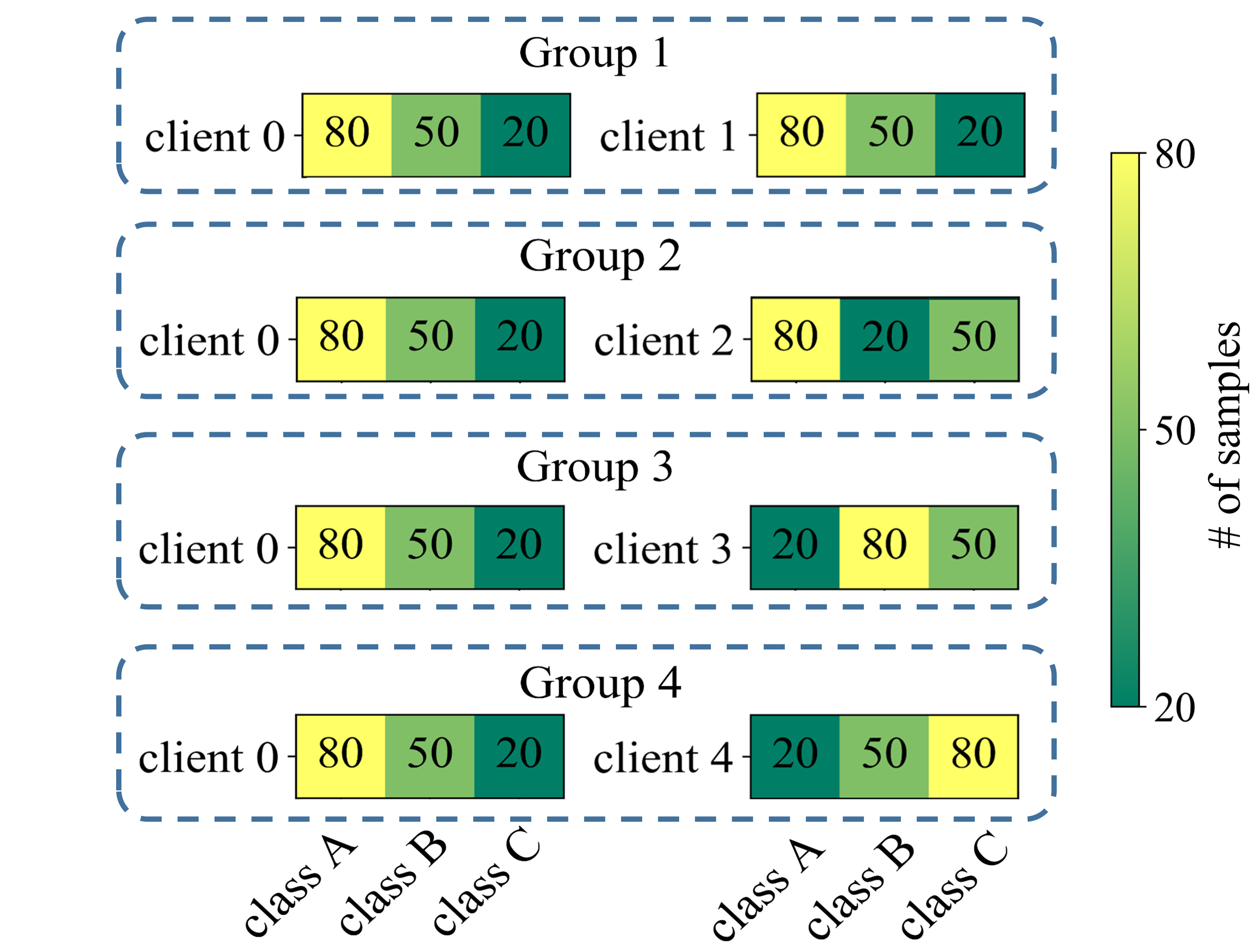}
\label{fig:dada}}
\hfil
\subfigure[Aggregate weight verse accuracy of FedAMP and DA-PFL.]{\includegraphics[scale=0.6]{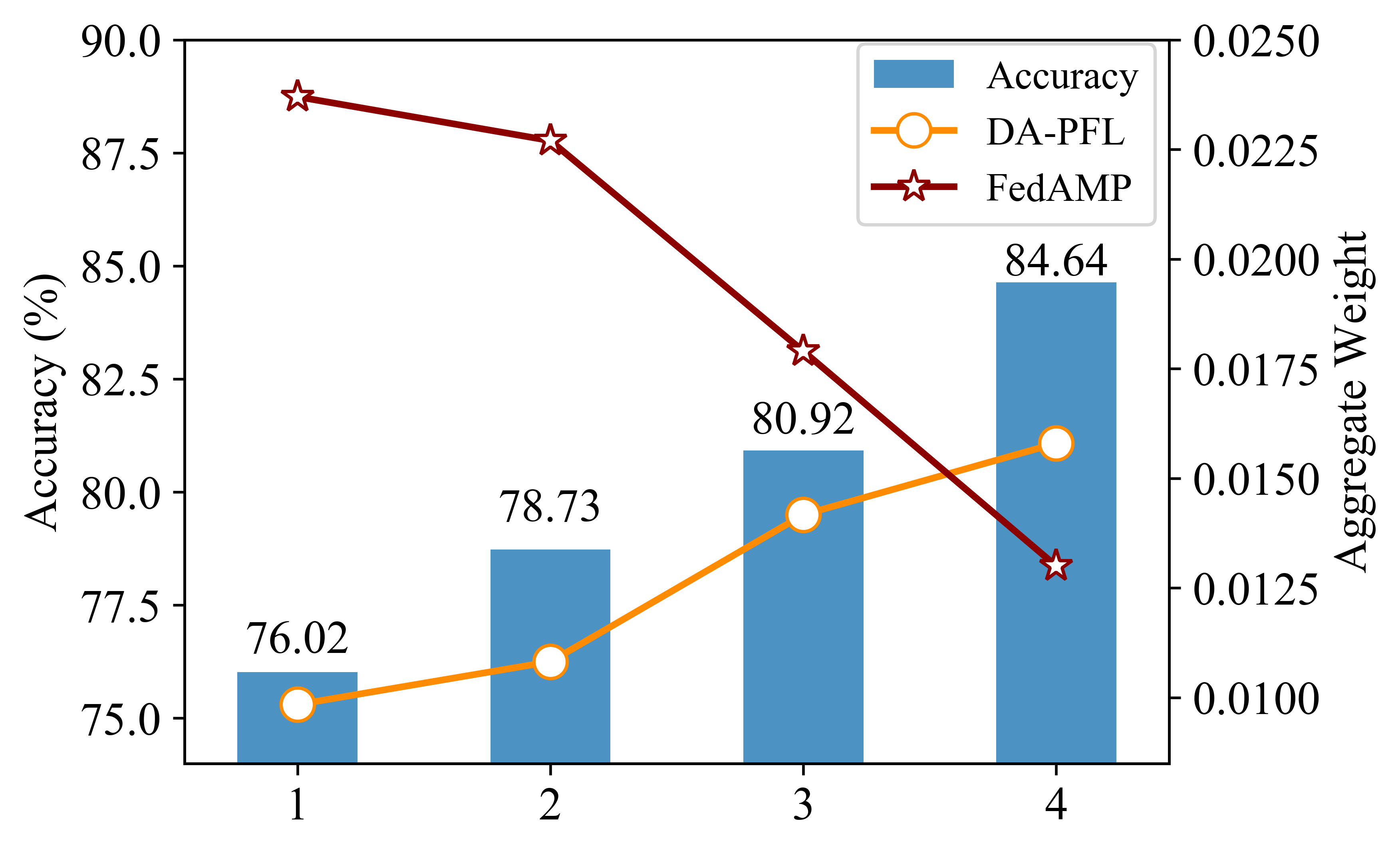}%
\label{fig:affinity}}
\caption{
The relationship between accuracy and weighting metrics, including the affinity of DA-PFL and the similarity of FedAMP. 
 }
\label{fig:observation}
\end{figure}


However, existing similarity-based aggregate strategies exacerbate the risk of class imbalance in PFL. As shown in Fig. \ref{fig:observation}, the FedAMP aggregation strategy is based on similarities between clients, while the proposed DA-PFL aggregation strategy is based on affinities between clients. 
Fig. \ref{fig:dada} indicates that group 1 has the highest similarity value between client 0 and client 1 because both clients contain the same number of samples (80, 50, and 20 samples) for the same class labels (class A, B, and C). On the other hand, group 4 has the lowest similarity value because client 0 and client 4 have different data distributions. The similarity-based aggregation strategy prefers to aggregate group 1, which leads to the sample number of Class A being much larger than the sample number of Class C and leads to an imbalance risk. On the other hand, we propose an affinity-based aggregation strategy that prefers to aggregate group 4 so that Class A's sample number is balanced with Class C's sample number. As shown in Fig. \ref{fig:affinity}, the red line denotes the aggregate weights of four groups in FedAMP. The higher aggregate weight means that clients have a higher priority to aggregate due to having more similar data distributions. We find that the highest aggregate weight of FedAMP is group 1 and the lowest aggregate weight is group 4, while the accuracy of group 1 is lower than that of group 4 (about an 8\% decrease). The above observation is because FedAMP cannot alleviate the class imbalance problem between Class A and Class C via aggregating similar clients. In contrast, aggregating client 0 and client 4 can balance the class distribution between Class A and Class C to increase the model accuracy. Inspired by the observation, we propose a novel personalized federated learning model called DA-PFL (orange line) to recalculate the aggregate weights of different groups. We find that the trend of the aggregate weight of DA-PFL is more similar to the trend of the model accuracy.

In this paper, we propose a new \textbf{D}ynamic \textbf{A}ffinity-based \textbf{P}ersonalized \textbf{F}ederated \textbf{L}earning model  \textbf{DA-PFL} 
to alleviate the class imbalanced problem. Concretely, we define an affinity metric that can effectively reflect the complementary relationship of class distribution between clients. We then design a new affinity-based aggregation strategy to assign higher aggregation weights to the more valuable clients with higher affinity. The experimental results indicate that the DA-PFL can outperform the state-of-the-art methods on real-world datasets. The contributions of this paper are summarized as follows:

\begin{itemize}
  \item We propose a novel dynamic affinity-based personalized federated learning model called DA-PFL to alleviate the class imbalanced problem. 
  
  \item We build an affinity metric to measure the affinity across clients from the view of the complementary class distribution to guide client aggregation. 
  
  \item We design a new dynamic aggregation strategy based on the affinity metric to dynamically select and aggregate valuable client models in each round to further improve personalized model performance. 
  
  \item We evaluate the DA-PFL model on three real-world datasets, and the results show that the DA-PFL can significantly improve the accuracies compared with the state-of-the-art methods.
\end{itemize}

\section{Related Work}
This section briefly reviews three personalized federated learning methods categories: regularization-based, architecture-based, and similarity-based.

\subsection{Regularization-based PFL}

Regularization-based methods aim to improve the optimization convergence and generalization ability of the personalized model from the perspective of model updating. For example, Li et al. \cite{li2021ditto} propose the ditto framework, which adds a regularization term in the client optimization objective to encourage the client model to approach the global model. Dinh et al. \cite{pfedme2020personalized} propose the pFedMe method, which adds $l_{2}$-norm regularization loss to balance the personalization and generalization ability of the client model. The above PFL methods mainly focus on heterogeneity problem \cite{yang2021regularized,kim2022multi,cheng2022differentially,karimireddy2020scaffold,li2019convergence} and ignore the class imbalance problem.
For the class imbalance problem, Fallah et al. \cite{perfedavgfallah2020personalized} propose the Per-FedAvg method, which obtains an initial global model by meta-learning in the early training round, and then trains the initial global model on local data of the client to obtain the final personalized model. 
Li et al. \cite{moonli2021model} propose MOON, which adds the model comparison loss to make each client model as close to the global model as possible and increase the distance from the historical client model.

\subsection{Architecture-based PFL}

Architecture-based methods aim to train personalized client models by decomposing the model parameters into shared and private layers. The client aggregates the share layers from other clients and combines their private layers into the personalized client model\cite{fedrepcollins2021exploiting,zhu2021federated,thapa2022splitfed,ma2022layer,li2021fedbn,Automatedtkde,ArchitectureSearchtkde}. Sun et al. \cite{sun2021partialfed} design the PartialFed method, where each client only loads partial layers of global model parameters instead of all parameters. The selection strategy for the layers is adaptively generated based on the parameters of the client model. Oh et al. \cite{oh2022fedbabu} propose a federated learning method called FedBABU, in which the classifier parameters of each client are without training and only fine-tuned at the end to achieve personalization. Chai et al. \cite{Chai_Ali_Zawad_Truex_Anwar_Baracaldo_Zhou_Ludwig_Yan_Cheng_2020} propose TiFL that selects the layers in real time over time based on accuracy. Some architecture-based PFL methods consider the class imbalance problem. For example, Collins et al. \cite{fedrepcollins2021exploiting} proposes a federated learning method called FedRep, which divides model parameters into feature extractor and classifier components. Each client retains its classifier parameters while sharing parameters of the feature extractor. The aggregation strategy for the feature extractor uses FedAvg \cite{mcmahan2017communication} across all clients. 
FedRep tries to solve the class imbalance problem with different class labels but the same number of samples. On the other hand, our work tries to solve the class imbalance problem with different class labels and different numbers of samples.

\begin{figure*}[t]
  \centering
  \includegraphics[scale=0.42]{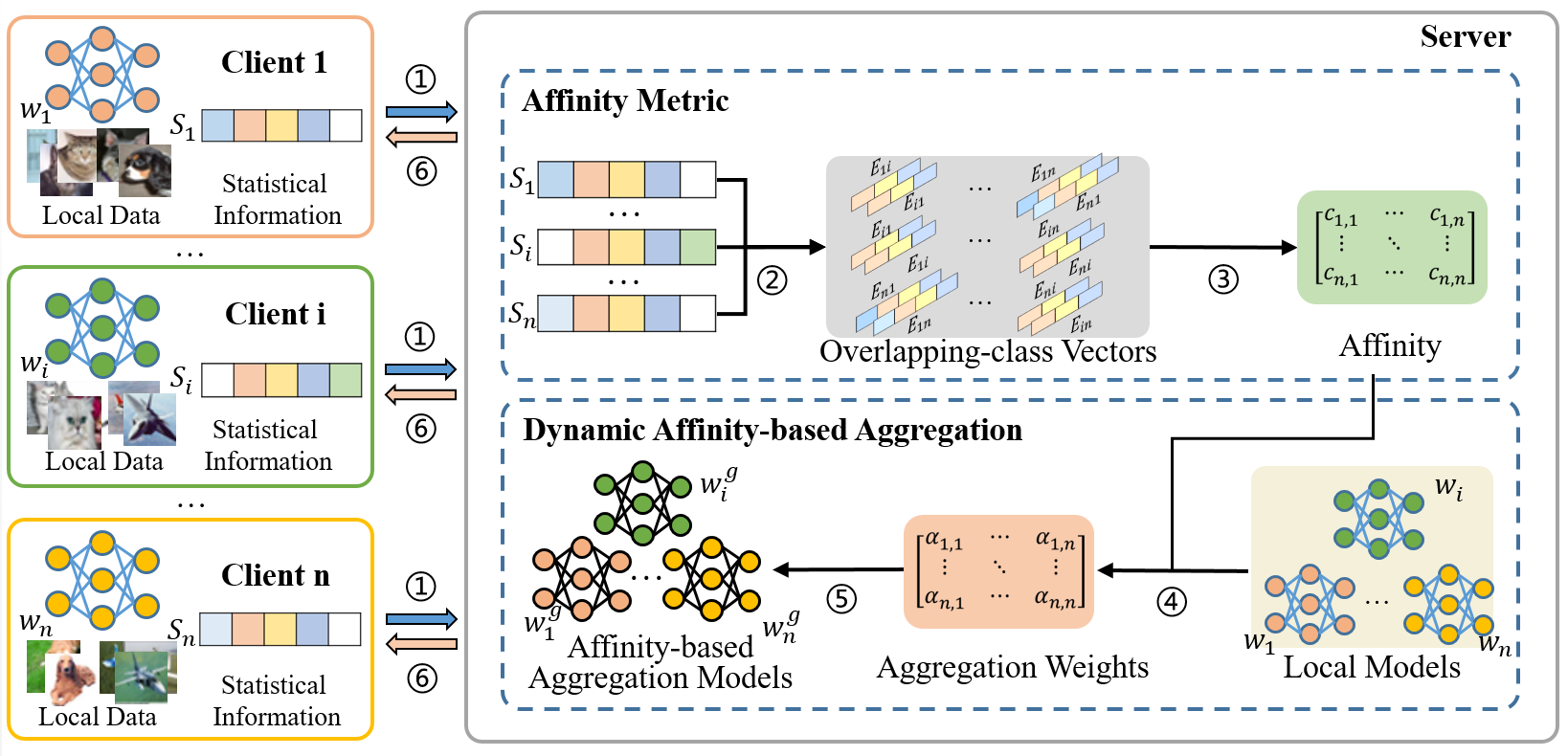}
  \caption{Illustration of the DA-PFL. 
  The workflow includes 6 steps: {\small{\textcircled{\scriptsize{\textbf{1}}}}\small}\@ clients send the statistical information and local models; {\small{\textcircled{\scriptsize{\textbf{2}}}}\small}, {\small{\textcircled{\scriptsize{\textbf{3}}}}\small}\@ the server calculates overlapping-class vectors and affinity for each client; {\small{\textcircled{\scriptsize{\textbf{4}}}}\small}, {\small{\textcircled{\scriptsize{\textbf{5}}}}\small}\@ the server calculates dynamic aggregated weights based on affinity and local models and aggregates affinity-based aggregation models for each client by dynamic aggregated weights; {\small{\textcircled{\scriptsize{\textbf{6}}}}\small}\@ clients download the affinity-based aggregation models and update their local models. 
 }
  \label{fig:2-method}
\end{figure*}

\subsection{Similarity-based PFL}
Similarity-based PFL methods aim to improve the performance of personalized models by encouraging clients with similar data distributions to collaborate\cite{sattler2020byzantine,liu2022deep,baek2023personalized,tu2021feddl,ktpflzhang2021parameterized,vo2022adaptive}. Clustering similar clients is one of the common similarity-based PFL methods, and it clusters clients into multiple groups to learn models for each group\cite{caldarola2021cluster,ChoCommunication,wang2022accelerating,ouyang2022clusterfl,fraboni2021clustered,song2023fast}. For example, Briggs et al. \cite{briggs2020federated} propose the FL+HC method that divides the client into several groups based on the similarity of their global model updates via hierarchical clustering technology. Duan et al.\cite{duan2021fedgroup} propose a FedGroup framework that uses the K-Means++ algorithm to group clients by the cosine similarity of the updated model between clients. Ruan et al. \cite{ruan2022fedsoft} propose the FedSoft method, which relaxed the restriction of the above hard clustering algorithms where a client belongs to only one clustering, allowing a client to belong to a mixture of multiple distributions. 
Ghosh et al. \cite{ifcaghosh2020efficient} propose an IFCA method that divides all clients into K disjoint groups according to their data distribution and iteratively minimizes the population loss function for each group. Another similarity-based method aggregates the client's model according to the similarity of the model parameters or data distribution. Liu et al.\cite{liu2021pfa} propose a PFA method that constructs a similarity matrix for each client using the Euclidean distance between private representations of local data. The PFA method uses the similarity matrix to complete group-wise aggregation. Similarly, Zhang et al. \cite{ktpflzhang2021parameterized} propose the KT-pFL method that linearly combines soft labels to improve cooperation between clients with similar data distributions during the distillation process. Huang et al.\cite{huang2021personalized} propose FedAMP, which calculates the personality weights between clients using attention-inducing functions and trains a personalized model for each client based on the personality weights. However, existing similarity-based methods still suffer the risk of class imbalance problems. In this work, we propose DA-PFL to alleviate the risk of class imbalance by introducing dynamic affinity collaboration between clients.

\section{DA-PFL Algorithm}

In this section, we propose a dynamic affinity-based personalized federated learning model to alleviate the class imbalanced problem. The DA-PFL workflow is illustrated in Fig.~\ref{fig:2-method}. Firstly, an affinity metric is designed to select clients with a complementary class distribution. Then, a dynamic affinity aggregation strategy is proposed to improve the performance of personalized models.

\subsection{Problem Definition}
We aim to collaboratively train a personalized model for each client with different class distributions.  
Given $N$ clients, and each with privacy data $D_1$, $D_2$,..., and $D_N$, respectively. 
$D=\cup _{i \in [N]} D_i $ is the whole dataset that contains $K$ classes. 
Let $ (\textbf{x}, y) \in {D_i} $ be a sample $\textbf{x}$ with label $y$ on client $i$. Federated learning allows each client to receive the global model from the server in each communication round. Each client optimizes the local model based on the global model and local private data with the following optimization objective: 
\begin{equation}
\min_{\boldsymbol{{w}_{i}}} \mathbb{E}_{(\boldsymbol{x}, y) \sim \mathcal{D}_{i}}\left[\mathcal{H}\left(\boldsymbol{{w}_{i}} ; \boldsymbol{{w}^{g}}, \boldsymbol{x}, y\right)\right],
\end{equation} 
where $\mathcal{H} $ is the loss function, $\boldsymbol{w_i}$ is the local model of client $i$ training with the local private data $D_i$. $\boldsymbol{w^{g}}$ is the global model that is aggregated by the local models of the last round.

We introduce a $L_2$-regularization term to minimize the divergence between the local and global models so that the local model can learn useful information from other clients. Therefore, the loss function for each client $i \in [N] $ is as follows:
\begin{equation}
\mathcal{H}\left(\boldsymbol{{w}_{i}} ; \boldsymbol{{w}_{i}^{g}}, \boldsymbol{x}, y\right):=\mathcal{L}\left(\boldsymbol{{w}_{i}} ;  \boldsymbol{x}, y\right)+ \frac{\lambda }{2}\left \| \boldsymbol{w_i} -\boldsymbol{w_i^g}\right \|^2,
\label{local_loss}
\end{equation}
where $\mathcal{L}$ is the cross-entropy loss function, $\lambda$ is a hyperparameter, $\boldsymbol{w_i}$ is the local model of client $i$, and $\boldsymbol{w_i^g}$ is the affinity-based aggregation model of client $i$ obtained in Section 3.3 according to affinity across clients.

\subsection{Affinity Metric}
We propose the affinity metric to accurately quantify the complementary relationship of clients according to statistical information \cite{tan2022fedproto,fedrodchen2021bridging,li2021fedrs}, such as the number of samples and the class index number. 
To calculate the affinity between each pair of clients based on complementary relationships, 
we create overlapping-class vectors based on client statistical information to reflect data correlation among clients. 
Let's the statistical information of client $i$ is $S_{i}$. 
The overlapping-class vectors $E_{ij}$ and $E_{ji}$ can be calculated as follows: 
\begin{equation}
\begin{aligned}
E_{ij}=\left\{S_{i,k} \mid  0 < k \le  K, S_{i,k}>0 \text{ and }  S_{j,k}>0 \right\}
\end{aligned}
\label{sij}
\end{equation}
\begin{equation}
\begin{aligned}
E_{ji}=\left\{S_{j,k} \mid  0 < k \le  K, S_{i,k}>0 \text{ and }  S_{j,k}>0 \right\}
\end{aligned}
\label{sji}
\end{equation}
where statistical information $S_{i,k}$ is the sample number of class $k$ in the client $i$. According to Eq.(\ref{sij}), if the client $i$ and $j$ both contain samples of class $k$, $E_{ij}$ will save the class number $S_{i,k}$. Similar to Eq.(\ref{sij}), the overlapping-class vector of client $j$ can be calculated by Eq.(\ref{sji}). The overlapping-class vector not only considers whether the class $k$ is overlapping, but the sample number of the class $k$ can reflect the degree of overlap between two clients.

Unlike the similarity-based aggregation strategy in FL, it tries to aggregate clients with similar data distributions, which leads to the risk of imbalance. We design an affinity metric to balance the sample number of classes based on overlapping-class vectors. The affinity between client $i$ and client $j$ is defined as follow:

\vspace{-0.2cm}
\begin{small}
\begin{equation}
\begin{aligned}
c_{i,j} = \frac{|E_{ij}|}{K} \! \cdot (2- \!\frac{ {\textstyle \sum_{\substack{e=0}}^{|E_{ij}|}} (S_{ij,e}-\! \overline s ) \cdot (S_{ji,e}- \! \overline s )}{\sqrt{ {\textstyle \sum_{\substack{e=0}}^{|E_{ij}|}}  (S_{ij,e}\!- \!\overline s )^2} \! \sqrt{ {\textstyle \sum_{\substack{e=0}}^{|E_{ij}|} (S_{ji,e} \!- \!\overline s )^2}} }),
\label{aff_e}
\end{aligned}
\end{equation}
\end{small}
and
\begin{equation}
\begin{aligned}
\overline s = \frac{{\textstyle \sum_{\substack{d=0}}^{|E_{ij}|}} S_{ij,e} + {\textstyle \sum_{\substack{d=0}}^{|E_{ij}|}} S_{ji,e}}{2|E_{ij}|},
\end{aligned}
\end{equation}
where $|E_{ij}|$ is the length of $E_{ij}$. $E_{ij,e}$ is $e$-th element of $E_{ij}$. 
$\overline s$ denotes the mean value of all elements of $E_{ij}$ and $E_{ji}$. 

In Eq.(\ref{aff_e}), we extend the adjusted cosine similarity \cite{ajcos} to calculate the affinity between clients. Extend adjusted cosine similarity utilizes the mean value of two overlapping label vectors $E_{ij}$ and $E_{ji}$ to mitigate the impact of significant differences in sample numbers, instead of their respective mean values. The ratio $\frac{|E_{ij}|}{K}$ controls the impact of the number of overlapping classes. We define affinity metric as negative extend adjusted cosine similarity and ensure its range to be positive. It should be noted that we set the affinities as the calculated minimum affinity between other clients when two clients have no overlap class in the statistical information. Subsequently, we normalize the obtained client affinity, restricting it to the range $c_{i,j} \in [0,1]$. 

\begin{algorithm}[t]
    \caption{DA-PFL Algorithm}
    \label{alg2}
    \textbf{Input}: Initial model $\boldsymbol{w_0}$; communication epoch $T$ \newline
    \textbf{Output}: Personalized models  $\boldsymbol{w_1},...,\boldsymbol{w_N}$ \\
    \vspace{-0.4cm}
    \begin{algorithmic}[1]
    \STATE \text{Clients upload their own statistical vector $S_{i}$}\\
    \STATE \textbf{Calculate} overlapping label vectors via Eq. (\ref{sij}) and (\ref{sji})\\
    \STATE \textbf{Calculate}  $c_{i,j}$ via Eq. (\ref{aff_e})\\
        \FOR{$t=1$ to $T$}
            \STATE \text{Server} randomly selects $m$ participating clients and  clients upload $\boldsymbol{w_1},...,\boldsymbol{w_m}$ \\
            \FOR{$i \in [m]$}
                \STATE \textbf{Calculate} $\theta_{i,j}$ via Eq. (\ref{e4})\\
                \STATE \textbf{Calculate} $\alpha_{i,j}$ via Eq. (\ref{e5})\\
                \STATE \textbf{Aggregate} $\boldsymbol{w_i^g} \gets  {\textstyle \sum_{j=1}^{m}} \alpha_{i,j}\boldsymbol{w_{j}}$ via Eq.(\ref{e6})
            \ENDFOR
            \STATE \textbf{Send} $\boldsymbol{w_i^g}$ to participating client $i \in [m]$\\
            \FOR{$i \in [m]$ \textbf{in parallel}} 
                \STATE \textbf{ClientUpdate}
            \ENDFOR
        \ENDFOR
        \STATE \textbf{ClientUpdate:} Update $\boldsymbol{w_i}^{t}$ via Eq. (\ref{wi_update})\\
    \end{algorithmic}
\end{algorithm}

\subsection{Dynamic Affinity-based Aggregation}

We propose dynamic aggregation that dynamically generates affinity-based aggregation models for clients based on the affinity between clients and client model parameters. For client $i$, we calculate the weight $\theta_{i,j}$ with client $j$ depending on the affinity $c_{i,j}$. The higher the affinity $c_{i,j}$, the higher aggregation weight $\theta_{i,j}$ with the client $j$. The calculation of the weight $\theta_{i,j}$ is formalized as follows:
\begin{equation}
\theta_{i,j}=\frac{c_{ij}}{ {\textstyle \sum_{j=1}^{m}}c_{ij} } (1-e^{-\frac{\left \| \boldsymbol{w_i}-\boldsymbol{w_j} \right \|^2+\varepsilon }{\sigma } }), i\ne j,
\label{e4}
\end{equation}
where $\boldsymbol{w_i}$ denotes the local model parameters of client $i$, and $\boldsymbol{w_j}$ denotes the model of other clients that participate in the aggregation of $\boldsymbol{w_i^g}$. $\sigma$ is a hyperparameter. $\varepsilon$ is a infinitesimal positive number.

The negative exponential term $(1-e^{-\frac{\left \| \boldsymbol{w_i}-\boldsymbol{w_j} \right \|^2+\varepsilon }{\sigma } })$ measures the nonlinear differences between different client model parameters. It enhances collaboration between complementary client models with differences. The exponential term is weighted by the affinity between clients, thus the aggregation model $\boldsymbol{w_{i}^{g}}$ for client $i$ will pay more attention to the client model with complementary data distribution. According to Eq. (\ref{e4}), the client can alleviate the risk of class imbalance by leveraging the knowledge contained in the complementary client models. It also helps clients adjust the aggregation targets and select useful clients for each training round based on model parameters. After the server calculates the weight $\theta_{i,j}$ for each pair of clients, the aggregation weight $\alpha_{i,j}$ can be calculated as follows: 
\begin{equation}
\alpha_{i,j}=\frac{\theta_{i,j}}{ {\textstyle \sum_{j=1}^{m}\theta_{i,j}} }.
\label{e5}
\end{equation}

In each training round, the server generates a unique affinity-based aggregation model for each client based on aggregation weights. For client $i$, the affinity-based aggregation model $\boldsymbol{w_i^g}$ can be calculated as follows: 
\begin{equation}
\boldsymbol{w_i^g} =\sum_{j=1}^{m} \alpha_{i,j} \boldsymbol{w_{j}}, i\ne j,  
\label{e6}
\end{equation}
where $\boldsymbol{w_j}$ is the local personalized model of client $j$. $m$ represents the number of clients who participated in training in this round of communication.

Once the server aggregates the affinity-based aggregation models $\boldsymbol{w_i^g}$ for all participating clients, it sends them to clients for training.  
The clients update the model parameters with local data under the guidance of the affinity-based aggregation model. According to Eq. (\ref{local_loss}), the update of the client model parameters is as follows: 
\begin{equation}
\begin{aligned}
\begin{small}
\boldsymbol{w_i}^t=\arg\min_{{\boldsymbol{w_i}^{t-1}}}\mathcal{L}\left(\boldsymbol{w}_{i}^{t-1};\boldsymbol{x}, y\right)+\frac{\lambda}{2}\left \| \boldsymbol{w_i}^{t-1}-\boldsymbol{w_i}^g\right \|^{2},
\end{small}
\label{wi_update}
\end{aligned}
\end{equation}
where $\boldsymbol{w_i}^t$ is the model parameter of client $i$ in round $t$ and $\lambda$ is the hyperparameter. 
Collaborating with affinity-based aggregation models enables client models to alleviate the risk of class imbalance and have better personalized performance. 
We summarize the steps of DA-PFL in Algorithm~\ref{alg2}. 

\begin{figure*}[h]
\centering
\subfigure[Low imbalance distribution, $\alpha$=1.0.]{
    \includegraphics[scale=0.42]{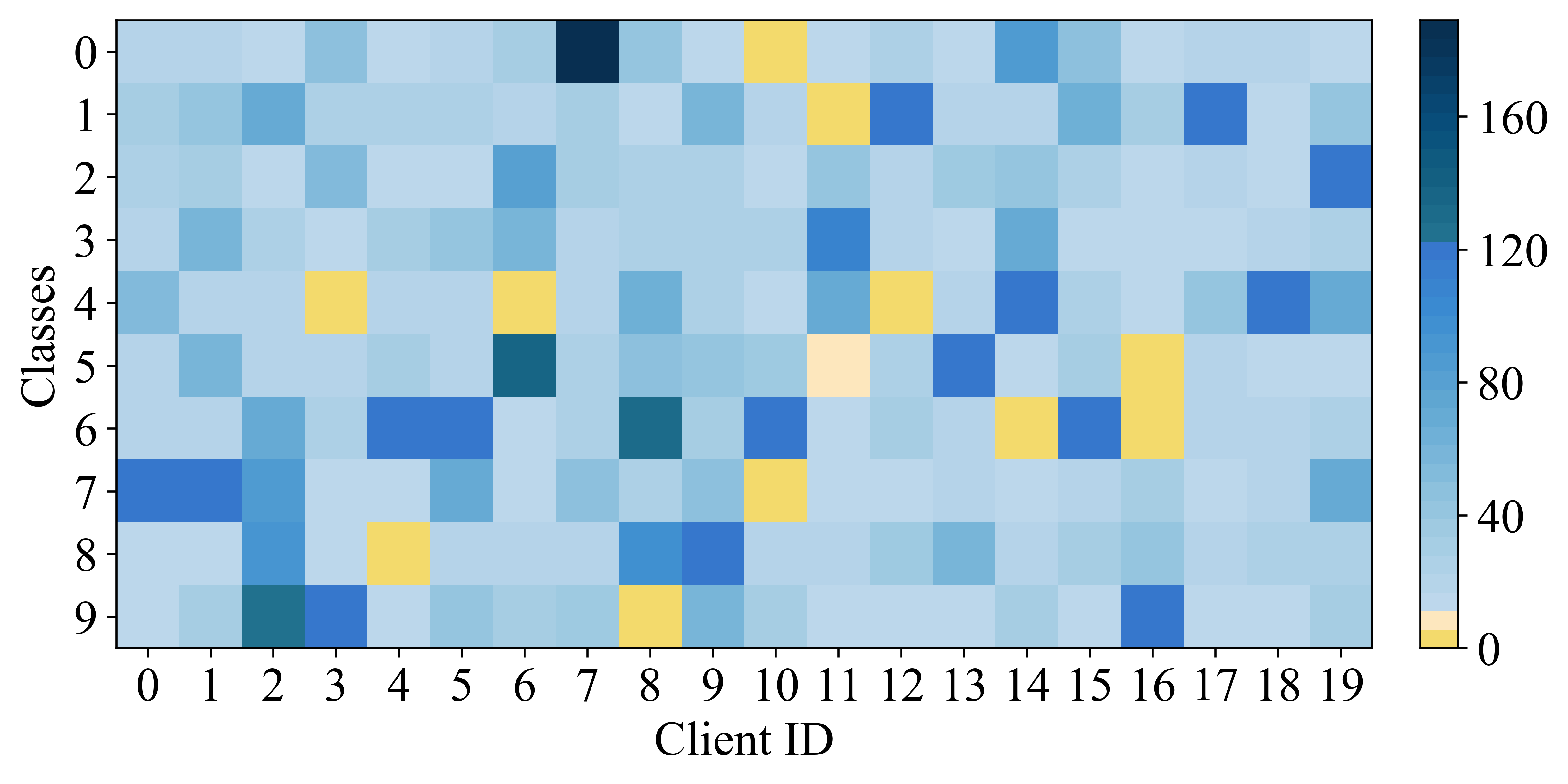}%
    \label{fig:short-a}}
\hspace{1.5mm}
\subfigure[High imbalance distribution, $\alpha$=0.5.]{
    \includegraphics[scale=0.42]{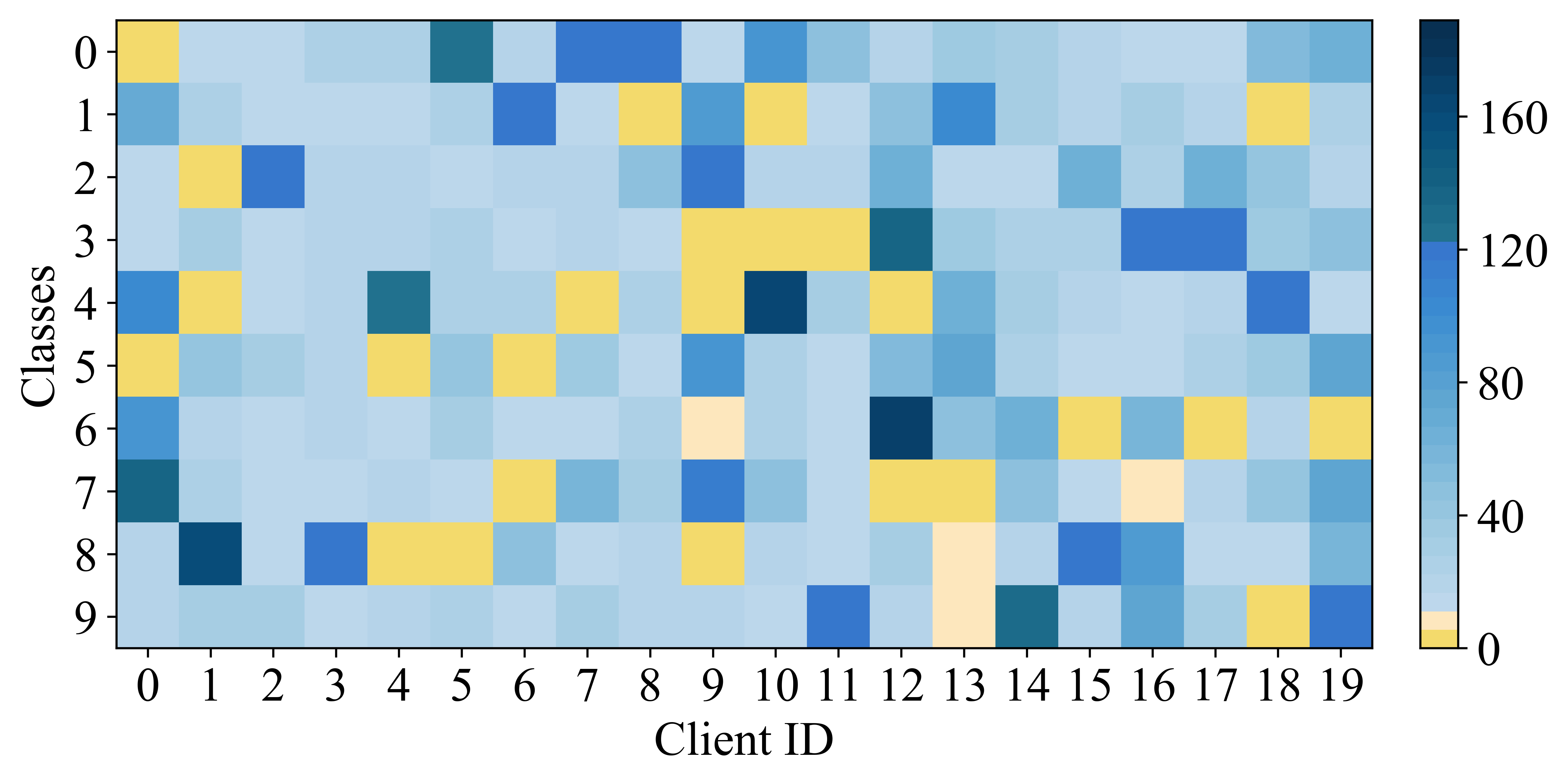}%
    \label{fig:short-b}}
\caption{Different imbalance distribution of class.}

\label{fig:short}
\end{figure*}

\section{Experiments Setting}
In this section, we introduce the experiment details of the proposed DA-PFL method and all the compared methods.

\subsection{Datasets}
Experiments are conducted on three widely used real-world FL benchmark datasets:
\begin{itemize}
\item{{\tt{CIFAR10}} \cite{krizhevsky2009learning}:  CIFAR10 is an image dataset in Federated Learning for classification tasks. It contains 60,000 color images from 10 different classes, each with a resolution of 32x32 pixels. This dataset is divided into 50,000 training images and 10,000 test images. In this paper, following \cite{fedrepcollins2021exploiting, shang2022federatedcreff}, we randomly allocate samples for each client in the training dataset, and select 10\% of the corresponding class samples in the test dataset as the test set for the client. }

\item{{\tt{CIFAR100}} \cite{krizhevsky2009learning}: 
This image dataset collects 100 classes of images, each class containing 500 training images and 100 test images. Each image is a 32 x 32 color image. 
Following the setting in \cite{achituve2021personalized}, we select the first 20 classes for training and testing.
We randomly assign training images to each client. The number of test images per client is 10\% of the training samples for each class.}

\item{{\tt{FEMNIST}} \cite{caldas2018leaf}: Federated Extended MNIST (FEMNIST) dataset is constructed by dividing Extended
MNIST \cite{cohen2017emnist} according to the writer of digits/characters. The dataset contains 805,263 images from 62 classes of 3550 users, with each image being 28 × 28 pixels. 
Following the previous work \cite{fedrepcollins2021exploiting}, we select 10 lowercase letters as the client dataset. We randomly select samples for each client, with training and testing sets accounting for 90\% and 10\%, respectively. }

\end{itemize}

\subsection{Class Imbalance Settings}
According to \cite{lin2020ensemble,shen2021agnostic,hsu2019measuringDir}, we also choose Dirichlet distribution to simulate the class imbalance environment for each dataset. We assume $q\sim Dir(\alpha) $ from a Dirichlet distribution, where $\alpha > 0 $ is a concentration parameter controlling the degree of imbalance among clients. 
According to the $\alpha$ of Dirichlet distribution, 
the training data classes for different clients are not exactly the same, and each class has a different number of images. 
The small $\alpha$ means that the client has a more heavily imbalanced class distribution, and vice versa. As show in Fig. \ref{fig:short}, we set two different class distributions including low imbalance ($\alpha=1$) and high imbalance ($\alpha=0.5$). 

Each column represents the classes owned by this client, and each row represents the distribution of samples in this class across all clients. 
For each cell, the darker blue color indicates a larger number of samples, while a yellow color indicates a small or non-existent number of samples. 
We can find that Fig. \ref{fig:short}(b) has a more heavily imbalance class distribution due to the more significant difference in sample numbers between different classes. 

\begin{table*}[t]
\caption{Average test accuracy under various imbalance settings in CIFAR10. 
The experiments in the table include 20 clients and 100 clients, with participation rates of 40\% and 20\%, respectively. Each client has two different unbalanced settings.}
\centering
\setlength\tabcolsep{17.5pt}
\begin{tabular}{cccccc}\toprule
\multicolumn{1}{l}{\multirow{2}{*}{\textbf{Family}}} & \multicolumn{1}{l}{\multirow{2}{*}{\textbf{Methods}}}  & \multicolumn{2}{c}{20 clients}  & \multicolumn{2}{c}{100 clients} \\ \cline{3-6} 
\multicolumn{2}{c}{} & \multicolumn{1}{c}{Low Imbalance} & \multicolumn{1}{c}{High Imbalance} & \multicolumn{1}{c}{Low Imbalance}   &  \multicolumn{1}{c}{High Imbalance} \rule{0pt}{8pt}  \\ \midrule 

\multicolumn{1}{l}{\multirow{1}{*}{Baseline}}  & \multicolumn{1}{l}{FedAvg}           & 48.93                    & 49.61                     & 47.46                        & 51.21            \\ \hline
\multicolumn{1}{l}{\multirow{2}{*}{\makecell{Regularization-based}}}  & \multicolumn{1}{l}{FedProx}             & 48.15                    & 49.83                      & 48.60                       & 51.88            \\ 
\multicolumn{1}{l}{} & \multicolumn{1}{l}{Ditto}             & 45.97                    & 48.78                      & 43.11                        & 48.00            \\ \hline
\multicolumn{1}{l}{\multirow{2}{*}{\makecell{Architecture-based}}} & \multicolumn{1}{l}{FedRep}            & 45.04                    & 47.87                      & 45.66                        & 48.78            \\  
\multicolumn{1}{l}{} & \multicolumn{1}{l}{CReFF}   & 52.68   & 50.87   & 55.85                        & 54.59            \\  \hline

\multicolumn{1}{l}{\multirow{4}{*}{\makecell{Similarity-based}}} & \multicolumn{1}{l}{IFCA}           & 49.27               & 48.44          & 44.77                     & 45.82            \\
\multicolumn{1}{l}{} & \multicolumn{1}{l}{FedPFA}           & 44.50                    & 38.42                      & 45.07                       & 38.23            \\
\multicolumn{1}{l}{} & \multicolumn{1}{l}{FedAMP}            & \underline{53.68}                     & 47.31                      & \underline{62.42}    & \underline{59.48}                  \\
\multicolumn{1}{l}{} & \multicolumn{1}{l}{FedSoft}           & 49.59                    & \underline{51.27}                      & 46.95                            & 49.88       \\  \hline

\multicolumn{1}{l}{\multirow{1}{*}{Proposed method}} & \multicolumn{1}{l}{DA-PFL}  & \makecell{\textbf{61.88} \\ (8.20$\uparrow$)}  & \makecell{\textbf{60.24} \\ (8.97$\uparrow$)}  & \makecell{\textbf{70.43} \\ (8.01$\uparrow$)}   & \makecell{\textbf{68.25} \\ (8.77$\uparrow$)}      \\ 
\bottomrule
\end{tabular}

\label{table:1}
\end{table*}

\subsection{Comparison Methods}
To better demonstrate the effectiveness of the proposed DA-PFL, we compare DA-PFL with nine state-of-the-art personalized federated learning methods, especially those that attempt to use the similarity of client data distributions to improve the performance of personalized models.
\begin{itemize}
\item{{\tt{FedAvg}}\cite{mcmahan2017communication}: FedAvg is the baseline that uses the weights calculated by the number of samples to aggregate local models of all clients.}
\item{{\tt{FedProx}}\cite{fedproxli2020federated}: FedProx is a regularization-based federated learning method that adds a proximal item to the loss function that improves the stability of the global model.}
\item{{\tt{Ditto}}\cite{li2021ditto}: Ditto is a regularized-based personalized federated learning method that can jointly solve personalized and global objectives alternatively. Ditto can learn personalized models for each client and incorporate a regularization term to alleviate the class imbalance risk.}
\item{{\tt{FedRep}}\cite{fedrepcollins2021exploiting}: Federated Representation Learning (FedRep) is an architecture-based personalized federated learning method, which splits the model into the low-dimensional representation layer and classifier layer to solve the class imbalance problem by sharing the feature representation.}
\item{{\tt{CReFF}}\cite{shang2022federatedcreff}: 
Classifier Re-training with Federated Features (CReFF) method is an architecture-based personalized federated learning method that retrains the classifier using the federated feature gradients and updated local model of clients. It solves the class imbalance problem by retraining the classifier.  
}
\item{{\tt{FedSoft}}\cite{ruan2022fedsoft}: FedSoft is a similarity-based personalized federated learning method that uses the broader similarities between the source cluster distributions but not the individual clients. FedSoft trains a personalized model for a client group with the knowledge of similar data distribution.}
\item{{\tt{FedPFA}}\cite{liu2021pfa}: Privacy-preserving Federated Adaptation (FedPFA) is a similarity-based personalized federated learning method that maintains a similarity matrix using sparsity to represent clients' raw data. 
FedPFA partitions the trained model of clients into different groups according to the similarity matrix to alleviate the risk of class imbalance, and each group shares a global model. }
\item{{\tt{IFCA}}\cite{ifcaghosh2020efficient}: Iterative Federated Clustering Algorithm (IFCA) is a similarity-based personalized federated learning framework that can identify the group membership of each client and optimize the personalized model for the group. IFCA also combines the weight-sharing technique and the alternating minimization algorithm to separate clients into distinct groups to alleviate the risk of class imbalance. }
\item{{\tt{FedAMP}}\cite{huang2021personalized}: Federated Attentive Message Passing (FedAMP) is a similarity-based personalized federated learning method that proposes to train a personalized cloud model for each client and use the message-passing mechanism to encourage pairwise collaboration with similar clients to alleviate the risk of class imbalance.}

\end{itemize}

\subsection{Implementation Details}
All experiments are run on the server with NVIDIA GeForce V100 GPU and Ubuntu 16.04 system and implemented based on PyTorch. For a fair comparison, we use the same hyperparameters for all methods. We adopt SGD as the optimizer for all optimization processes. For the CIFAR10 and FEMINST datasets, we set the learning rate to 0.01. And the learning rate in CIFAR100 datasets is 0.001. The batch size for experiments on these three datasets is set to 10. For the CIFAR10 and CIFAR100 datasets, we follow \cite{fedrepcollins2021exploiting,caldas2018leaf} and use the two different CNN models. These CNN models are composed of two convolutional layers and three fully connected layers, but each layer has different scales. 
For the FEMNIST dataset, we follow\cite{li2019feddane,caldas2018leaf} to use the MLP with two hidden layers. We show the average accuracy of the client models on the local test dataset for the last 10 rounds as the method accuracy. 

In addition, for regularization-based methods, we use the same coefficients of the regularization term. 
For architecture-based methods, we divided the model architecture based on the experimental settings in \cite{fedrepcollins2021exploiting}.
For similarity-based clustering methods, we set the number of clusters at 3 and 7 based on different client numbers (20 and 100 clients), respectively. 

 \section{Results Analysis}
We compare DA-PFL with the nine popular methods under different numbers of clients and different imbalance distributions. Concretely, we set the number of clients to 20 and 100 with a participation rate of 0.4 and 0.2 under High imbalance distribution ($\alpha$=0.5) and Low imbalance distribution ($\alpha$=1), respectively. 

\subsection{Performance of DA-PFL}
In this subsection, we report the average accuracy of all personalized models. Results highlight the advantages of the DA-PFL method over state-of-the-art methods.

\paragraph{Result on CIFAR10} 
Table~\ref{table:1} reports the performance in the CIFAR10 dataset under various client quantities and imbalance distributions. The proposed DA-PFL achieves the best accuracy among all client quantities and imbalance distributions. When the client is 20, and the class distribution is highly imbalanced, the accuracy of the DA-PFL test improved by 8.97\% compared to the FedSoft method, which is the best comparison method. We set up 100 clients and verify the effect of the DA-PFL with different numbers of participating clients under the same class imbalance distribution. When the class distribution is low imbalance, compared to the FedAMP method, our method improves up to 8.01\%. The accuracy of the DA-PFL method is also improved by 8.77\% when the class distribution is high imbalance. The result indicates that DA-PFL has the optimal result in different imbalance distributions and different numbers of clients.
\begin{table*}[!ht]
\caption{Average test accuracy under various imbalance settings in CIFAR100. The experiments in the table include 20 clients and 100 clients, with participation rates of 40\% and 20\%, respectively. Each client has two different unbalanced settings.} 
\centering
\setlength\tabcolsep{17.5pt}
\begin{tabular}{cccccc}\toprule
\multicolumn{1}{l}{\multirow{2}{*}{\textbf{Family}}} & \multicolumn{1}{l}{\multirow{2}{*}{\textbf{Methods}}}  & \multicolumn{2}{c}{CIFAR100, 20 clients}  & \multicolumn{2}{c}{CIFAR100, 100 clients} \\ \cline{3-6} 
\multicolumn{2}{c}{} & \multicolumn{1}{c}{Low Imbalance} & \multicolumn{1}{c}{High Imbalance} & \multicolumn{1}{c}{Low Imbalance}   &  \multicolumn{1}{c}{High Imbalance} \rule{0pt}{8pt}  \\ \midrule

\multicolumn{1}{l}{\multirow{1}{*}{Baseline}}  & \multicolumn{1}{l}{FedAvg}           & 38.57                    & 41.78                     & 38.01                        & \underline{40.73}            \\ \hline
\multicolumn{1}{l}{\multirow{2}{*}{\makecell{Regularization-based}}}  & \multicolumn{1}{l}{FedProx}             & 38.68                    & \underline{41.84}                      & 38.22                       & 40.18            \\ 
\multicolumn{1}{l}{} & \multicolumn{1}{l}{Ditto}             & \underline{41.33}                    & 40.32                      & 36.31          & 34.74            \\ \hline
\multicolumn{1}{l}{\multirow{2}{*}{\makecell{Architecture-based}}} & \multicolumn{1}{l}{FedRep}            & 36.22                    & 39.69                      & 35.32                        & 37.36            \\  
\multicolumn{1}{l}{} & \multicolumn{1}{l}{CReFF}   & 37.47     & 40.09    & 32.84       & 33.60             \\  \hline

\multicolumn{1}{l}{\multirow{4}{*}{\makecell{Similarity-based}}} & \multicolumn{1}{l}{IFCA}           & 26.01               & 27.85         & 20.94                     & 21.01            \\
\multicolumn{1}{l}{} & \multicolumn{1}{l}{FedPFA}           & 26.62                    & 23.92                      & 25.58               & 23.22           \\
\multicolumn{1}{l}{} & \multicolumn{1}{l}{FedAMP}            & 36.65           & 32.35              & \underline{42.19}    & 39.94                  \\
\multicolumn{1}{l}{} & \multicolumn{1}{l}{FedSoft}           & 31.05                    & 33.80            & 25.07           & 28.78       \\ \hline

\multicolumn{1}{l}{\multirow{1}{*}{Proposed method}} & \multicolumn{1}{l}{DA-PFL}  & \makecell{\textbf{57.20} \\ (15.87$\uparrow$)}  & \makecell{\textbf{58.28} \\ (16.44$\uparrow$)}  & \makecell{\textbf{68.49} \\ (26.30$\uparrow$)}   & \makecell{\textbf{66.81} \\ (26.08$\uparrow$)}      \\ 
\bottomrule
\end{tabular}

\label{table:2}
\end{table*}

\begin{table*}[ht]
\caption{Average test accuracy under various imbalance settings in FEMNIST. The experiments in the table include 20 clients and 100 clients, with participation rates of 40\% and 20\%, respectively. Each client has two different unbalanced settings.}
\centering
\setlength\tabcolsep{17.5pt}
\begin{tabular}{cccccc}\toprule

\multicolumn{1}{l}{\multirow{2}{*}{\textbf{Family}}} & \multicolumn{1}{l}{\multirow{2}{*}{\textbf{Methods}}}  & \multicolumn{2}{c}{FEMNIST, 20 clients}  & \multicolumn{2}{c}{FEMNIST, 100 clients} \\ \cline{3-6} 
\multicolumn{2}{c}{} & \multicolumn{1}{c}{Low Imbalance} & \multicolumn{1}{c}{High Imbalance} & \multicolumn{1}{c}{Low Imbalance}   &  \multicolumn{1}{c}{High Imbalance}  \rule{0pt}{8pt} \\ \midrule 

\multicolumn{1}{l}{\multirow{1}{*}{Baseline}}  & \multicolumn{1}{l}{FedAvg}           & 66.83                    & \underline{68.16}                     & 67.40                        & \underline{71.65}            \\ \hline
\multicolumn{1}{l}{\multirow{2}{*}{\makecell{Regularization-based}}}  & \multicolumn{1}{l}{FedProx}             & \underline{68.00}          & 28.62                      & 30.33                       & 32.42            \\ 
\multicolumn{1}{l}{} & \multicolumn{1}{l}{Ditto}             & 64.91       & 64.30                      & \underline{69.78}          & 65.11           \\ \hline
\multicolumn{1}{l}{\multirow{2}{*}{\makecell{Architecture-based}}} & \multicolumn{1}{l}{FedRep}            & 33.50 & 34.96 & 31.11 & 37.76           \\  
\multicolumn{1}{l}{} & \multicolumn{1}{l}{CReFF}   & 55.23     & 52.56     & 61.98     & 51.59      \\  \hline

\multicolumn{1}{l}{\multirow{4}{*}{\makecell{Similarity-based}}} & \multicolumn{1}{l}{IFCA}           & 59.06 & 50.12 & 24.35 & 40.81            \\
\multicolumn{1}{l}{} & \multicolumn{1}{l}{FedPFA}  & 53.43 & 54.57 & 59.83 & 62.97  \\
\multicolumn{1}{l}{} & \multicolumn{1}{l}{FedAMP} & 62.13 & 64.11 & 48.33 & 55.71 \\
\multicolumn{1}{l}{} & \multicolumn{1}{l}{FedSoft}   & 29.84 & 28.62 & 27.13 & 33.42   \\ \hline
\multicolumn{1}{l}{\multirow{1}{*}{Proposed method}} & \multicolumn{1}{l}{DA-PFL}  & \makecell{\textbf{74.43} \\ (6.43$\uparrow$)}  & \makecell{\textbf{76.17} \\ (8.01$\uparrow$)}  & \makecell{\textbf{80.04} \\ (10.26$\uparrow$)}   & \makecell{\textbf{79.34} \\ (7.69$\uparrow$)}      \\ 
\bottomrule
\end{tabular}

\label{table:3}
\end{table*}

\paragraph{Result on CIFAR100} Table~\ref{table:2} reports the performance in the CIFAR100 dataset under different client numbers and imbalance distributions. When the number of clients is 20 and the class distribution is high imbalance, FedProx is the optimal result in the comparison methods. And the accuracy of DA-PFL is higher than that of FedProx, up to 16.44\%. When we set the client quantity to 100, the accuracy of DA-PFL is significantly improved. When the class distribution is low imbalance, compared to the FedAMP method, the accuracy of DA-PFL improves up to 26.30\%. Furthermore, high imbalance distribution will not significantly reduce the accuracy of DA-PFL, which means that DA-PFL has a good performance when solving the class imbalance problem. However, similarity-based methods significantly decrease model performance due to ignoring the risk of class imbalance distribution.

\paragraph{Result on FEMNIST} Table~\ref{table:3} shows the performance in the FEMNIST dataset under various client numbers and imbalance distributions. DA-PFL is still optimal among all the comparison methods. When the number of clients is 20 and the class distribution is low imbalance, the average accuracy of DA-PFL is higher than IFCA and obtains an improvement of 15.37\%. When the class distribution is high imbalance, the accuracy of IFCA decreases significantly. Furthermore, we set the client number to 100, and the average accuracy of the proposed DA-PFL method still outperforms all compared methods. When the class distribution is low imbalance, compared to the Ditto method, the accuracy of DA-PFL improves up to 10.26\%. The accuracy gap between the DA-PFL and the optimal comparison methods increases when the class distribution is high imbalance. It indicates that the compared methods ignore the risk of class imbalance distribution, and the DA-PFL method is unaffected by the class imbalance distribution. 

\subsection{Accuracy Analysis of Various Classes}
In the CIFAR100 dataset, we selected 20 clients and set the distribution as high imbalance to further analyze the improvement effect of the DA-PFL in the various classes. We first divide the classes into Few (less than 30 samples), Medium (30-80 samples), and Many (more than 80 samples) classes according to the number of samples, and then calculate their accuracy, respectively. The results are shown in Table~\ref{table:4}. The regularization-based and architecture-based methods do not consider the effect of sample number on the model's performance. Few and Medium Class accuracy is very low due to a few samples, resulting in poor model performance. In the FedRep method, the accuracy of Few Class is only 9.95\%, which is much lower than the 43.30\% of DA-PFL. 

\begin{table*}[t]
\caption{The test accuracy of various classes in the CIFAR100. The number of clients is 20, with participation rates of 40\%. }
\centering
\setlength\tabcolsep{26pt}
\begin{tabular}{cccccc}\toprule
\multicolumn{1}{l}{\multirow{2}{*}{\textbf{Family}}} & \multicolumn{1}{l}{\multirow{2}{*}{\textbf{Methods}}} & \multicolumn{4}{c}{CIFAR100, 20 clients} \\ \cline{3-6} \rule{0pt}{8pt}
    &            & Many          & Medium           & Few     & Avg       \\  \midrule 
\multicolumn{1}{l}{\multirow{1}{*}{Baseline}}  & \multicolumn{1}{l}{FedAvg}       &59.90 &25.13 &13.26 &41.78      \\ \hline
\multicolumn{1}{l}{\multirow{2}{*}{\makecell{Regularization-based}}}  & \multicolumn{1}{l}{FedProx}   &59.08 &25.92 &9.61 &\underline{41.84}     \\
 \multicolumn{1}{l}{} & \multicolumn{1}{l}{Ditto}      &\underline{60.06} &22.53 &7.69 &40.32       \\ \hline
\multicolumn{1}{l}{\multirow{2}{*}{\makecell{Architecture-based}}} & \multicolumn{1}{l}{FedRep}    &58.29 &22.76 &9.95 &39.69     \\
\multicolumn{1}{l}{} & \multicolumn{1}{l}{CReFF}      &41.08 &\underline{40.38} &\underline{30.12} &40.09    \\ \hline
\multicolumn{1}{l}{\multirow{4}{*}{\makecell{Similarity-based}}} & \multicolumn{1}{l}{IFCA}  &30.58 &22.27 &14.17 &27.85          \\
\multicolumn{1}{l}{} & \multicolumn{1}{l}{FedPFA}       &24.62 &24.45 &19.92 &23.92          \\
\multicolumn{1}{l}{} & \multicolumn{1}{l}{FedAMP}      &34.63 &33.47 &25.03 &32.35           \\
\multicolumn{1}{l}{} & \multicolumn{1}{l}{FedSoft}      &59.04 &8.14 &1.14 &33.80             \\ \hline
\multicolumn{1}{l}{\multirow{1}{*}{Proposed}} & \multicolumn{1}{l}{DA-PFL}                     &\makecell{\textbf{60.31} \\ (0.25$\uparrow$)}          &\makecell{\textbf{58.52} \\ (18.14$\uparrow$)}        &\makecell{\textbf{43.30} \\ (13.18$\uparrow$)}    &\makecell{\textbf{58.28} \\ (16.44$\uparrow$)}     \\ 
\bottomrule
\end{tabular}

\label{table:4}
\end{table*}

Furthermore, the accuracy of the Few Class on the similarity-based method is also very low, with the FedSoft method at a minimum of 1.14\%. Similarity-based methods provide higher weights for clients with similar data distributions. The personalized model fits well on the Many Class while ignoring the Few Class. 
DA-PFL not only maintains the accuracy of the Many Class but also greatly improves the accuracy of the Few classes. 
DA-PFL selects clients with complementary class distribution through high affinity to supply the knowledge of Few Class since the personalized model obtains full training in Few Class, significantly improving the accuracy of Few Class.

\subsection{Ablation Study}

In this subsection, we conduct ablation experiments to investigate the effect of each component of DA-PFL in CIFAR-10. 
The results are shown in Table \ref{table:6}, DA-PFL-V1 without affinity metric and dynamic aggregation components, the accuracy is the lowest, only 45.97\%. 
DA-PFL-V2 only uses the affinity metric component, resulting in a 10.24\% increase compared to DA-PFL-V1.   
The results indicate that the affinity metric component can effectively find and aggregate other client models with appropriate affinity weights. 
The dynamic aggregation component is based on the affinity metric component, so we don't design experiments with only the dynamic aggregation component. 
The affinity metric and dynamic aggregation components make up the proposed DA-PFL. 
Compared to DA-PFL-V2, the accuracy of DA-PFL still increased by 15.91\%. 
The result proves that the proposed dynamic aggregation component can effectively adjust the aggregation weights of the clients for each personalized global model and address the risk of class imbalance. 

\begin{table}[t]
\caption{Ablation Study of DA-PFL in CIFAR10 dataset with 20 clients, imbalanced degree $\alpha$ = 1.0.}
\begin{center}
\centering
\setlength\tabcolsep{7.5pt}
\begin{tabular}{ccccc}
\toprule
Methods   & \makecell{Affinity\\Metric}  & \makecell{Dynamic\\Aggregation}   &Accuracy(\%)  &$\uparrow$ \\ 
\midrule 
DA-PFL-V1     & \ding{53}   & \ding{53}  &45.97 & -     \\ 
DA-PFL-V2    &$\checkmark$ & \ding{53}  &56.21 &+10.24     \\ 
DA-PFL     &$\checkmark$ &$\checkmark$  &61.88 &+15.91  \\
\bottomrule
\end{tabular}

\label{table:6}
\end{center}
\end{table}

\subsection{Convergence Speed of DA-PFL}
\begin{figure*}[t]
  \centering
  \includegraphics[scale=0.31]{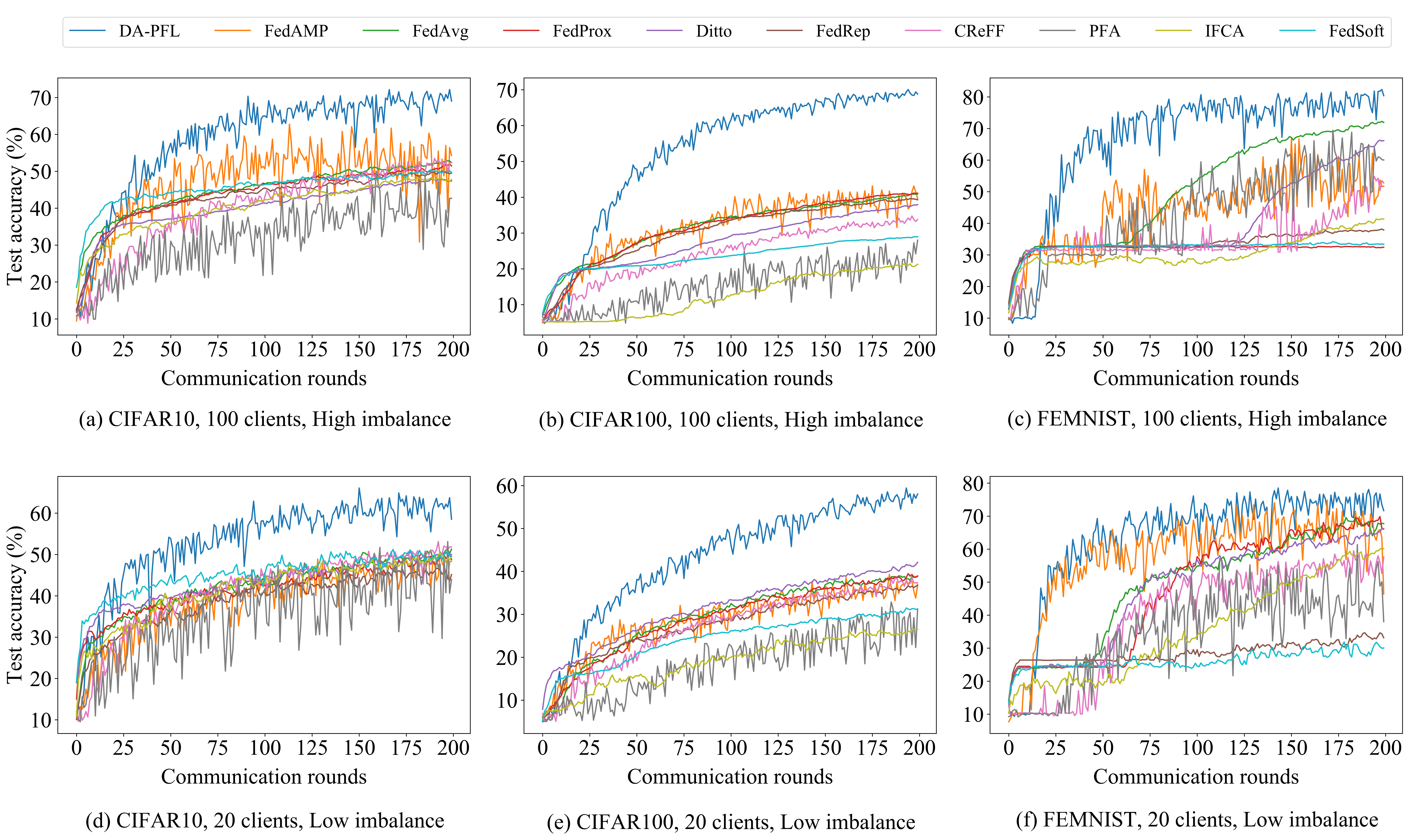}
  \caption{Accuracy of different communication rounds on test data set in model training.}
  \label{fig:4}
\end{figure*}

The convergence process of all methods under different imbalance distributions and datasets is shown in Fig.~\ref{fig:4}. In the CIFAR10, Fig.~\ref{fig:4}(a) and Fig.~\ref{fig:4}(d) show that the convergence speed of DA-PFL is significantly faster than all comparison methods in different imbalance distributions, Fig.~\ref{fig:4}(b) and Fig.~\ref{fig:4}(e) also present the same trend in CIFAR100. In FEMINIST, Fig.~\ref{fig:4}(c) and Fig.~\ref{fig:4}(f) show that the convergence of the PFA and FedAMP methods is unstable because both methods exacerbate the risk of class imbalance by dividing clients into different groups with similarity. The accuracy of FedSoft and FedRep is much lower than other methods and performs poorly in convergence because they cannot adapt to the heavy class imbalance distribution.

\begin{table}[t]
\caption{Communication rounds to reach target test accuracy ($C_{acc}(x)$) for High imbalance distribution, 20 clients.}
\centering
\Large
\resizebox{\linewidth}{!}{ 
\begin{tabular}{ccccccc}
\toprule
\multicolumn{1}{c}{\multirow{2}{*}{Methods}} & \multicolumn{2}{c}{CIFAR10}  & \multicolumn{2}{c}{CIFAR100} & \multicolumn{2}{c}{FEMINST} \rule{0pt}{8pt} \\ \cline{2-7} 
&$C_{acc}(40\%)$   &$C_{ac}(50\%)$   & $C_{acc}(20\%)$  & $C_{ac}(30\%)$    & $C_{acc}(40\%)$ & $C_{acc}(50\%)$  \\ 
\midrule
FedAvg     &59  &\underline{153}  &36   &79  &57  &\underline{74}  \\ 
FedProx    &62  &184 &36   &91  &74 &89   \\ 
Ditto     &\underline{45} &178  &25   &74  &63 &\underline{74} \\
CReFF     &56 &156  &45 &105  &62  &83    \\ 
FedPFA    &73 &159  &83 &181  &69 &98     \\ 
FedAMP    &69 &194  &\underline{24} &\underline{74}  &\underline{19} &\textbf{22}   \\ 
\hline  \rule{0pt}{8pt}
DA-PFL     &\textbf{17}  &\textbf{35}  &\textbf{17} &\textbf{28}  &\textbf{18}  &\textbf{22}  \\
\bottomrule
\end{tabular}}
\label{table:5}
\end{table}

Following the previous works \cite{fedetcho2022heterogeneous,dafkd,fedres}, Table \ref{table:5} shows the communication round required by DA-PFL and comparison methods to reach target test accuracy in three datasets. 
The best results are highlighted in bold. 
We can find that the convergence speed of DA-PFL is much faster than of the compared method, and use fewer communication rounds to achieve the target accuracy. 
The result indicates that DA-PFL can effectively select required client models by affinity and aggregate them with appropriate weights to alleviate class imbalance, thereby accelerating the convergence speed.

\section{Conclusion}
In this paper, we find that similarity-based federated learning ignores the risk of imbalance data distribution, resulting in poor performance of the personalized model. Therefore, we proposed DA-PFL, a new dynamic affinity-based personalized federated learning to alleviate risk. We first proposed affinity metrics to measure the complementary relationship of data distribution across the client. We then design a dynamic aggregation strategy based on affinity metrics to select useful clients and assign aggregation weights, which can improve the performance of the client's personalized model. Experimental results on real-world datasets show that the performance of our method exceeds the state-of-the-art methods.

\section*{Acknowledgment}
This work was supported by the National Natural Science Foundation of China No.62076079 and the Guangdong Major Project of Basic and Applied Basic Research NO.2019B030302002.







\end{document}